\documentclass[man,floatsintext,donotrepeattitle]{apa7}  
\usepackage[american]{babel}
\usepackage{csquotes}
\usepackage[style=apa,backend=biber,natbib=true]{biblatex}

\usepackage{float}
\usepackage{placeins}

\usepackage{graphicx}
\graphicspath{{imgs/}}
\usepackage{subcaption}

\usepackage{tabu}
\usepackage{booktabs}

\usepackage{tikz}
\usepackage{amsmath}
\usepackage{amsthm}

\usepackage{mathtools}
\usepackage{verbatim}

\usepackage[flushleft]{threeparttable}

\usepackage[ruled,vlined]{algorithm2e}
\usepackage{soul}
\usepackage{hyperref}

\usepackage{threeparttable}

\usepackage{etoolbox}
\DeclarePairedDelimiterX{\abs}[1]{\lvert}{\rvert}{\ifblank{#1}{{}\cdot{}}{#1}}

\newcommand{\ie}{\textit{i}.\textit{e}., }
\newcommand{\eg}{\textit{e}.\textit{g}. }

\makeatletter
\newcommand*{\rom}[1]{\expandafter\@slowromancap\romannumeral #1@}
\makeatother

\newcommand{\comm}[1]{}

\title{Resolving Anomalies in the Behaviour\\
of a Modularity Inducing Problem Domain\\
with Distributional Fitness Evaluation}
\shorttitle{Distributional Fitness Evaluation}
\authorsnames{Zhenyue~Qin,Tom~Gedeon,R.I.~(Bob)~McKay}
\authorsaffiliations{{School of Computing, the Australian National University\\
Canberra ACT 2600 Australia}}
\authornote{\addORCIDlink{Zhenyue Qin, zhenyue.qin@anu.edu.au}{0000-0002-3471-6280}: corresponding author\\
\addORCIDlink{Tom Gedeon,  tom@cs.anu.edu.au}{0000-0001-8356-4909}\\
\addORCIDlink{Bob McKay, rimanucs@gmail.com}{0000-0003-3160-5822}}
\leftheader{Qin,Gedeon,McKay}

\abstract{Discrete gene regulatory networks (GRNs) play a vital role in the study of robustness and modularity. A common method of evaluating the robustness of GRNs is to measure their ability to regulate a set of perturbed gene activation patterns back to their unperturbed forms. Usually, perturbations are obtained by collecting random samples produced by a predefined distribution of gene activation patterns. This sampling method introduces stochasticity, in turn inducing dynamicity. This dynamicity is imposed on top of an already complex fitness landscape. So where sampling is used, it is important to understand which effects arise from the structure of the fitness landscape, and which arise from the dynamicity imposed on it. Stochasticity of the fitness function also causes difficulties in reproducibility and in post-experimental analyses. 

We develop a deterministic distributional fitness evaluation by considering the complete distribution of gene activity patterns, so as to avoid stochasticity in fitness assessment. This fitness evaluation facilitates repeatability. Its determinism permits us to ascertain theoretical bounds on the  fitness, and thus to identify whether the algorithm has reached a global optimum. It enables us to differentiate the effects of the problem domain from those of the noisy fitness evaluation, and thus to resolve two remaining anomalies in the behaviour of the problem domain of~\citet{espinosa2010specialization}. We also reveal some properties of solution GRNs that lead them to be robust and modular, leading to a deeper understanding of the nature of the problem domain. We conclude by discussing potential directions toward simulating and understanding the emergence of modularity in larger, more complex domains, which is key both to generating more useful modular solutions, and to understanding the ubiquity of modularity in biological systems.}


\begin{document}
\maketitle

\section{Introduction}
\label{sec:intro}
This paper extends the research of~\citet{10.1162/isal_a_00166}, which investigated anomalies in the modularity emergence model of~\citet{wagner1996does} and~\citet{espinosa2010specialization}, providing deeper understanding of the anomalous evolutionary behaviours noted in that paper. It does so by using distributional methods. The analysis is based directly on consideration of properties of the distribution underlying the sampling based fitness evaluation of~\citet{espinosa2010specialization}. Interestingly, for the size 10 patterns that formed the bulk of that work, it is computationally reasonable to directly evaluate the distribution and use it as an alternative fitness function. Doing so permits deeper insight into the behaviour of the system, and reveals that most of the important evolutionary properties -- especially the emergence of modularity -- result directly from the distribution, and do not require the presence of stochasticity in fitness evaluation arising from sampling noise. Conversely, some of the anomalous behaviours we found in~\citet{10.1162/isal_a_00166} do result from that stochasticity, and do not appear in runs using distributionally-determined fitness.

Modular systems arise everywhere in biology, at all scales -- in ecosystems, in physical organisation (organs, vascular systems, connective tissues etc.), in peripheral and central nervous systems, in intra-organ structure, between intracellular organelles, and at the genetic level, in gene regulatory networks (GRNS). Of these, GRNs are the only modular system that extends right across the known biological evolutionary spectrum (eukaryota, bacteria, archaea, viruses). Thus, the choice of GRNs as the underlying representation for A. Wagner's model seems prescient. However this occurrence right across the spectrum, with some specific GRNs appearing in the last common ancestor of all today's biological organisms, makes it very difficult to directly observe, or infer much about, their original emergence. Thus simulation is almost the only option to gain fuller understanding.

In artificial evolutionary systems, it is difficult to directly define modularity as a system goal without also predefining the structure, negating the creative freedom of the system. Thus it is also inportant to understand how it emerges in nature without being explicitly targeted, and how the same emergence may be encouraged in artificial systems. 

In the next section, we discuss the three main theories in the next section,
explaining why Espinosa-Soto and A. Wagner’s (2010) theory is particularly relevant in understanding the biological emergence of modularity.

Robustness is a fundamental property underlying all levels of organisations within biological systems~\citep{kitano2002systems}. It enables a system to maintain its functionalities despite facing internal and external perturbations~\citep{kitano2004biological}.
Taking networks as examples, biological systems and processes have evolved the capacity to quickly adapt to constantly varying environments and to be robust to  failures caused by both internal and external errors~\citep{binitha2012survey}. 

Studies of robustness commonly randomly sample a set of perturbed patterns, and test whether the model can regulate the disturbed patterns back to their original forms~\citep{wang1994robustness}. Topics have included  Boolean networks~\citep{aracena2009robustness}, Hopfield networks~\citep{hopfield1984neurons}, gene regulatory networks~\citep{espinosa2010specialization} among many others. Although stochastic sampling is realistic in mimicking the non-deterministic processes of nature, it causes problems for theoretical and experimental analysis:
\begin{enumerate}
\item{Repeatability:} sampling stochasticity means that the same input can lead to different outputs in repetitions of trials. 
\item{Computational Cost:} some evolutionarily relevant processes occur only once in a lifetime (\eg being eaten by a predator), so are inexpensive to simulate; for others (heartbeats in a human life $\sim10^{10}$; synapse firing a few orders of magnitude more), accurate simulation, even on present-day computers, is infeasible.
\item{Identifying Optima:} understanding the distribution of local optima is a key goal of evolutionary analysis, but sampling stochasticity makes it very difficult to determine whether a particular point is a true local optimum, or whether its apparent high fitness is by chance.
\item{Separating Fitness Landscape and Stochastic Effects:} some evolutionary behaviours may result from the underlying structure of the fitness landscape, others from the stochasticity of fitness evaluation; full understanding requires differentiating the two, but this is infeasible with purely stochastic methods.

\end{enumerate}

In this paper, we show the possibility of leveraging the entire distribution to analyse robustness of discrete system biological models in a maximally deterministic manner. We justify its feasibility from two perspectives: 
\begin{enumerate}
	\item Distributions from which the discrete model perturbations are sampled are analytically tractable, and can also be computationally feasible. For example, the sampling process of \citet{espinosa2010specialization} follows an underlying binomial distribution over the perturbations of the target, so the size of the fitness landscape for a target size of $N$ is the number of perturbations, \ie $2^N$; in their experiments $N \le 15$, hence feasible to compute directly. 
	\item When $N$ is too large to manage, we may still directly measure high probability events from the distribution, and sample only the low-probability portion. This  significantly reduces the number of events to tackle, while minimising stochasticity and maximally employing information from the distribution. 
\end{enumerate}

In summary, we outline our contributions as follows: 
\begin{enumerate}
	\item Using distributional evaluation, we are able to determine that the emergence of modularity in the model of~\citet{espinosa2010specialization} is a property of the distribution itself, not of the dynamicity arising from stochasticity. 
	\item Conversely, we are able to determine that some of the anomalous properties of the stochastic model do arise from that dynamicity.
	\item More generally, we exhibit a new avenue for studying robustness and related properties, leveraging the entire perturbation distribution in place of traditional sampling-based methods.
\end{enumerate}

\section{Related Work and Motivation}
\label{sec:rel_work_moti}
In this section, we present a brief overview of the emergence of modularity, of GRNs and of the GRN Model of~\citet{espinosa2010specialization}. Then we introduce the motivation of this paper. 

The main competing theories to explain the emergence of modularity in~\citet{wagner2007road} are those of~\citet{kashtan2005spontaneous} that modularity arises from modularly-varying evolutionary goals; of~\citet{clune2013evolutionary} that modularity is attributable it to biological parsimony pressures; 
and of Espinosa-Soto and A. Wagner (2010) that modularity is a by-product of  evolutionary specialisation.
Judging by citations, the model of~\citet{clune2013evolutionary} has recently been dominant. There is little doubt of its importance for evolving modular engineering systems, but its relevance to biological evolution is less obvious. Parsimony pressures have long been studied in genetic programming~\citep{doi:10.1162/evco.1995.3.1.17}, yet are notoriously difficult to tune -- too strong a pressure (relative to the primary objective) and one is left with tiny but highly unfit solutions; too weak, and complexity runs riot. In biology, as Clune et al. argue,  modularity is ubiquitous. Yet there is no obvious mechanism to tune the many different parsimony pressures required to explain its widespread emergence. Clune et al. avoid the need to specify a trade-off through use of the highly engineered NSGA-II multi-objective evolutionary algorithm~\citep{996017}, which is perfectly fine for engineering modularity in artificial systems, but questionable as a model of evolution based on natural selection (NSGA-II uses population-wide computations that would require a 'hidden hand' in biological systems). Thus, at least in understanding biology, it is worth re-examining alternative explanations. 

\citet{espinosa2010specialization} traced the emergence of modularity to specialisation. Their work is based on A. Wagner's evolutionary GRN model, which has witnessed wide application in a variety of computational biological studies~\citep{bergman2003evolutionary,azevedo2006sexual,huerta2007wagner}.
It is important to acknowledge that their system also included a parsimony mechanism, but in the mutation operator, not as a second objective. This parsimony mechanism is not sufficient on its own to generate modularity, specialisation is required; equally important, the parsimony of the mutation operator requires little tuning, a wide range of values suffice.

\subsection{Gene Regulatory Network (GRN)}
A GRN is a collection of molecular regulators that coordinate interactions between genes (including both the protein-coding DNA sequences and regulatory non-coding DNA sequences), RNAs and proteins~\citep{xiao2009tutorial}. GRNs are central to the operation of all known forms of cellular life (eukaryota, bacteria and archaea~\citep{10.1093/bfgp/elp056}) and viruses~\citep{Bensussen2018}.
The network structure of GRNs demonstrates a high level of modularity, considered to be a key contributor to robustness~\citep{kirsten2011evolution}. 

The evolutionary discrete GRN model of~\citet{espinosa2010specialization} is based on the discrete Boolean network GRN model, which traces its origins back to the work of~\citet{KAUFFMAN1969437}. The model is highly abstracted, and thus cannot model the detailed behaviour of real GRNs, which act in continuous time, with continuous levels of promotion and repression, and potentially continuous states. As Kauffman argues, the abstraction is nevertheless sufficient to model important aspects of biological GRNs, in particular their temporal behaviour, and to do so with much lower computational cost, and with a more readily analysable system, than more detailed abstractions. This is particularly so when determining whether specific behaviours are achievable by GRNs: if a discrete GRN exhibits a specific behaviour, then ipso facto, it can be performed by GRNs -- although the reverse may not  necessarily be the case (\ie continuous GRNs may exhibit behaviours unachievable by discrete GRNs).  In our case, if modularity emerges in discrete GRNs, then the same mechanism should also be able to generate modularity in the more general setting of continuous systems. The computational cost arguments are all the more powerful when we model the behaviour, not of a single fixed GRN as in Kauffman's work, but the evolution of many GRNs over a timescale of many generations. We would not be able to find the computational resources to model a comparable continuous system; and even more important, we can see no way to conduct an equivalent of the distributional model that we analyse in much of the rest of this paper.

\subsection{Espinsa-Soto and Wagner's GRN Model}
\label{subsec:esw_grn_model}
\begin{figure}[htb]
	\centering
	\includegraphics[width=\linewidth]{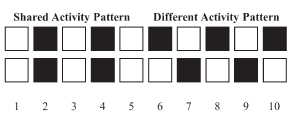}
	\caption{A target consisting of two gene activation patterns, where white and black squares represent active ($+1$) and inactive ($-1$) genes}
	\label{fig:gap}
\end{figure}

The model of~\citet{espinosa2010specialization} abstracts cellular homeostasis, in which a cell can recover from small perturbations to a target state and recover that state. It comprises two components: one or more targets, and the GRN itself. A target is a set of gene activation patterns, 
represented by a vector of $N$ binary values, with $+1$ and $-1$ respectively representing activity and inactivity, 
so that there can be $2^{N}$ targets overall. Figure \ref{fig:gap} depicts two target activation patterns consisting of $10$ genes that happen to have a modular structure of two modules of five genes each; the activation in the first module of the patterns is identical (shared), but opposite in the second (divergent).  

The gene state is regulated by a GRN $g$, which controls the activation pathway of the organism. For a pattern of N genes, it is abstracted as a ternary $N^2$ transition matrix $g = \lbrack g_{ji} \rbrack$ with entries over $\lbrace -1, 0, 1 \rbrace$, representing repression, independence or activation of gene $i$ by gene $j$. A gene activity pattern regulated by this network is a Boolean row vector $s = [s^1,...,s^N]$. 
The state transition is modelled by:
\begin{equation}
\label{eq:grn}
A(g,s)  =\sigma[g.s]
\end{equation}
where $\sigma(x)=1$ if $x>0$, $\sigma(x)=-1$ otherwise (applied elementwise). 

\begin{figure}[htbp]
	\centering
	\includegraphics[width=0.45\linewidth,keepaspectratio]{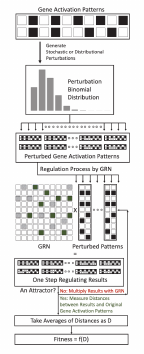}
	\caption{Flow Chart of the Fitness Evaluation in the~\citet{espinosa2010specialization} GRN model. Dark untextured squares represent activation, grey texture repression and white lack of influence. }
	\label{fig:esw_model}
\end{figure}

The model focuses on the evolution of the $N \times N$ GRN matrix, generally by an evolutionary algorithm. This can lead to a terminological confusion. In the modeled biology, there are $N$ genes in the activation pattern; but considered at the evolutionary algorithm level, the evolving chromosome consists of $N^2$ genes. Where we need to distinguish these, we refer to the former as a ``pattern gene'', and the latter as a ``network gene'' or ``network node''. Figure \ref{fig:esw_model} presents a flow chart of the fitness evaluation in the model.  

In the rest of this paper, we will concentrate on the work of Espinosa-Soto \& A. Wagner (2010) with the activation patterns of figure~\ref{fig:gap}, where an evolutionary system is evolved first for 500 generations with a target consisting solely of the first activation pattern, and then is evolved for the remainder of the time (1500 further generations) with both activation patterns as target.

\subsection{Anomalous Behaviours of the GRN Model and Our Motivation}
\label{subsec:anomalies}
\begin{figure}[htb]
	\centering
	\begin{subfigure}[b]{0.48\linewidth}
		\includegraphics[width=\linewidth]{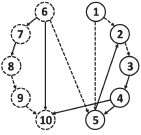}
		\caption{Before Removal}
	\end{subfigure}
	\begin{subfigure}[b]{0.48\linewidth}
		\includegraphics[width=\linewidth]{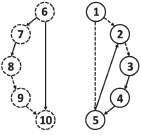}
		\caption{After Removal}
	\end{subfigure}
	\caption{Illustration of deterministically removing all the inter-module connections of a GRN. Solid and dashed circles represent different modules. Solid and dashed arrows stand for activation and repression respectively. }
	\label{fig:manual_removing_edge}
\end{figure}

\begin{figure}[htb]
	\centering
		\includegraphics[width=\linewidth]{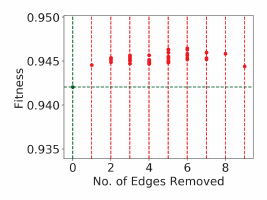}
	\caption{Fitness increases while removing inter-module connections from a fit GRN. Green dot represent the initial fitnesses, and red dots show the resulting fitnesses from inter-module connection removal. }
	\label{fig:fitness_increase_path}
\end{figure}

Our initial work on this GRN model under typical genetic algorithm settings revealed a number of anomalies~\citep{10.1162/isal_a_00166}. In summary, despite relatively fit, modular GRNs emerging in simulated evolutions, they could often be readily improved in both fitness and modularity by manually removing all inter-module connections, as in Figure~\ref{fig:manual_removing_edge}. Yet evolutionary search does not find these improvements, despite mutation biases appearing to favour finding them. Figure~\ref{fig:fitness_increase_path} reveals that this  is not due to discontinuous gradients: starting with the most robust/fittest GRN from the final generation  of a typical run and removing non-modular edges one-by-one reveals a path of steadily improving fitness to a fully modular GRN. 

\begin{table}[htbp]
	\centering
	\caption{Numbers of Unrecoverable Elementarty Perturbations by Weight.}
	\label{tb:prob_not_back}
	\begin{tabu} to \linewidth {X[1]X[4]X[5]} 
		\toprule
		Weight & No. of Perturbations & Unrecoverable\\
		\midrule
		0 & $\binom{10}{0}$ & $0$ \\ 
		\midrule
		1 & $\binom{10}{1}$  & $0$ \\ 
		\midrule
		2 & $\binom{10}{2}$  & $0$ \\ 
		\midrule		
		3 & $\binom{10}{3}$  & $\binom{5}{3}$ \\ 
		\midrule		
		4 & $\binom{10}{4}$  & $ \binom{5}{3} \cdot \binom{5}{1} + \binom{5}{4} $ \\ 
		\midrule		
		5 & $\binom{10}{5}$  & $ \binom{5}{3} \cdot \binom{5}{2} + \binom{5}{4} \cdot \binom{5}{1} + \binom{5}{5} $ \\ 
		\midrule		
		6 & $\binom{10}{6}$  & $ \binom{5}{3} \cdot \binom{5}{3} + \binom{5}{4} \cdot \binom{5}{2} + \binom{5}{5} \cdot \binom{5}{1} $ \\ 
		\midrule		
		7 & $\binom{10}{7}$  & $ \binom{5}{3} \cdot \binom{5}{4} + \binom{5}{4} \cdot \binom{5}{3} + \binom{5}{5} \cdot \binom{5}{2} $ \\ 
		\midrule		
		8 & $\binom{10}{8}$  & $\binom{10}{8}$ \\ 
		\midrule		
		9 & $\binom{10}{9}$  & $\binom{10}{9}$ \\ 
		\midrule		
		10 & $\binom{10}{10}$  & $\binom{10}{10}$ \\ 
		\bottomrule
	\end{tabu}
\end{table} 

\begin{table}[htbp]
	\centering
	\caption{Gene Activity Patterns (Target)}
	\label{table:genactiv}
	\begin{tabu} to \linewidth {X[2]X}  
		\toprule
		Target & Introduction\\ 
		Pattern & Stage\\
		\midrule
		+1 -1 +1 -1 +1 -1 +1 -1 +1 -1 & 0 \\ 
		+1 -1 +1 -1 +1 +1 -1 +1 -1 +1 & 500 \\
		\bottomrule
	\end{tabu}
\end{table}

\begin{table}[htbp]
	\centering
	\caption{Parameters of the Evolutionary Simulations}
	\label{table:simparams}
	\begin{tabu} to \linewidth {XXX}  
		\toprule
		Pattern Size & GRN Size & Initial Density\\
		10 & 100 & 0.2\\
		\midrule
		\texttt{\#} Perturbations & Perturbation Rate  & Population Size\\ 
		$2^N$ or 500 & 0.15 & 100\\
		\midrule
		Mutation Rate & Activation Rate & Crossover Rate \\ 
		0.2 &0.5  & 0.2 \\
		\midrule
		Crossover Type & Selection (Size) & Replication Rate\\
		Diagonal & Tournament (3) & 0\\
		\midrule
		Max. Generation & Trials/Treatment & Significance Test \\
		2000 & 100 & Mann Whitney  \\
		\bottomrule
	\end{tabu}
\end{table}

\begin{table}[htbp]
	\centering
	\caption{ Mean Best Fitnesses and Modularity Q Values of Fittest GRNs (over 100 runs) from the Final Generation, for Distributional vs Stochastic Fitness Evaluation}
	\label{table:dist_stoc_fit}
	\begin{tabu} to \linewidth {X[2.4]X[1.5]X[1.5]X[1.5]X[1.5]X[3.5]} 
		\toprule
		& \multicolumn{2}{l}{Distributional} & \multicolumn{2}{l}{Stochastic} & p-value \\
		&	Mean	&	SD	&	Mean	&	SD	&	\\
		\midrule 
		Fitness & 0.9395 & 0.0183 & 0.9370 & 0.0202 & $1.04 \times 10^{-12}$\\
		\multicolumn{3}{l}{Distributional Equivalent} & 0.9256 & &\\
		\midrule
		Modularity 	& 0.9785 & 0.2840 & 0.9513 & 0.3011 & 0.5016 \\
		\bottomrule
	\end{tabu}
\end{table}

\section{Distributional Fitness Evaluation}
\label{sec:dist_fit_eval}
We hypothesised that the anomalies discussed in the preceding section might arise from the stochasticity of the sampling process of~\citet{espinosa2010specialization}: that in a population converged close to a local optimum, using order-based (tournament) selection, small stochastic variations in fitness might make it difficult to follow weak gradients. To evaluate this hypothesis, we need to separate the effects of the underlying fitness landscape from the effects of stochastic sampling. Fortunately, this is not hard to do, both in principle, and in this case, in practice. 

In common with other GRN robustness models, for example~\citep{siegal2002waddington, masel2004genetic, ciliberti2007robustness}, \citet{espinosa2010specialization} sample perturbations 
stochastically, then study the recovery of the original pattern. They
use a binomial model: 500 perturbations of the locations in the pattern are identically and independently sampled, with a probability of being perturbed of $p=0.15$, the recovery by each GRN generates a reward based on the level of recovery, then the reward is averaged over the sampled perturbations. Thus we can compute the expected fitness of a GRN by tracing its behaviour over all 1024 possible perturbations, and weighting appropriately. This produces a deterministic fitness metric, and at a computational cost $1024/500$ (i.e. roughly double) that of~\citet{espinosa2010specialization}. We call this method distributional fitness evaluation. The underlying idea is extensible beyond discrete GRNs to a wide range of computational studies of discrete networks, including other genetic networks~\citep{huang1999gene}, Boolean neural networks~\citep{bhide1993boolean}, and Hopfield networks~\citep{gurney2014introduction}. 

\subsection{Definition and Advantages of Distributional Fitness Evaluation}
Partially following the ideas of~\citet{espinosa2010specialization}, we can extend the (one step) action $A(g,s)$ of an $N \times N$ GRN $g$ on an activation state $s$ of length $N$ of equation~\ref{eq:grn} to its recursive application as
\begin{equation*}
\begin{aligned}
A^0(g,s) &= s\\
A^{t+1}(g,s) &= A(g,A^t(g,s)) 
\end{aligned}
\end{equation*}

We define an elementary perturbation $e$ of length $N$ as a vector of $\lbrace -1, 1 \rbrace$, so that a perturbation of a target state $s$ (also of length $N$) in the sense of~\citet{espinosa2010specialization} is the pairwise product $e \odot s$. Following Boolean usage, the weight $w(e)$ of an elementary perturbation is the number of $-1$ values.

\citet{espinosa2010specialization} followed the regulatory process for $t_0=20$ steps:
\begin{equation}
\label{eq:regulation}
G(g, e \odot s) = 
\begin{cases}
s &\text{if }A^t(g,e \odot s)=s \text{  for } t<t_0 \\
A^{t_0}(g,e \odot s) & \text{otherwise}
\end{cases}
\end{equation}
and used two auxiliary functions to weight contributions:

\begin{equation*}
\begin{aligned}
f(g) &= 1-e^{(-3 \cdot g)} \\
\gamma(x) &= (1 - x)^{5}
\end{aligned}
\end{equation*}

Putting it together, we evaluate the effectiveness of GRN $g$ in recovering state $s$ as 
\begin{equation}
\label{eq:distributionalfitness}
\begin{aligned}
&F(g,s) = \\
&f \bigg( \sum_{n=0}^N p_n \cdot \frac{1}{\abs{E_n}} \sum_{e \in E_n} \gamma \bigg( H \big( G(g, e \odot s),s  \big) \bigg) \bigg)
\end{aligned}
\end{equation}
where $E_n$ is the set of elementary perturbations of length $N$ and weight $n$, $p_n$ is the probability $p_n \sim B(N, p)$ from the binomial distribution, and $H$ is the Hamming distance.

Compared with the random sampling of perturbations of~\citet{espinosa2010specialization}, distributional fitness evaluation offers the following advantages: 
\begin{enumerate}
	\item Determinism: evaluating the fitness of a GRN multiple times will give the same fitness each time, while preserving essentially the same fitness landscape as sampling. This allows us to disentangle the effects of the fitness landscape itself, and that of stochastic noise.
	\item Global Optimum Analysis: we can determine bounds on the fitness, and test whether those bounds are achieved; with a stochastically evaluated fitness, it is infeasible to determine whether further improvement is possible.
	\item Speed Optimisation: Due to the determinism of fitness evaluation, cacheing of previously computed fitness may reduce wasted computation.
\end{enumerate}

\subsection{Upper Bounds on Distributional Fitness}
\label{subsubsec:corollary}
An obvious upper bound for the fitness of a GRN in the scenario of ~\citet{espinosa2010specialization} is for the GRN to return all perturbations of a target to the corresponding target. When there is only a single target, this is attainable, and for the $10 \times 10$ GRNs that are the focus of this study, is readily evolved. However we follow~\citet{espinosa2010specialization} in using a two-stage evolution, in the second stage of which there are two targets. In this case, this upper bound is not attainable. It is easy to see why. Consider the first target of Figure~\ref{fig:gap}. If it is perturbed by an elementary perturbation of weight 0 (i.e. it is unperturbed), then we would expect a GRN to readily recover it (it isn't required to do anything). But consider the second target perturbed by an elementary perturbation whose first five locations are 1, with the last five being -1. The resulting perturbed target is identical to the previous one. Hence the GRN must map it to the same end result: which is a Hamming distance of 5 from its target state. In fact, for every perturbation of target 1, there is a perturbation of target 2 that gives the same starting state for the GRN. At most one of them can be returned to its target by the GRN. Which choice is best?

To answer this question, consider the binomial probabilities of elementary perturbations by weight: to two decimal places: $<0.20, 0.35, 0.28, 0.13, 0.04, 0.01, 0.00, 0.00, 0.00, 0.00, 0.00>$. So perturbations of weights 0, 1 and 2 carry the most influence in equation~\ref{eq:distributionalfitness}, and for higher weights the influence decreases monotonically. Now for two elementary perturbations to conflict in this way, they must be identical in the first five places (since the two targets are identical there), and inverse on the last five (since the two targets are inverse there). So the choice is easy: the elementary perturbation with the least weight (therefore 0, 1 or 2) in the second half should be mapped back to the corresponding target. In this case, the Hamming distance between the regulated and original patterns is $0$, and the result of function $\gamma$ in Formula \ref{eq:distributionalfitness} is $1$. Conversely, for second-half weights of 3, 4 or 5, the GRN will regulate the pattern to the opposite form. In such cases, the Hamming distance is $5$ and $\gamma$ returns $0.03125$. 

In summary, specialising equation~\ref{eq:distributionalfitness} to the targets $s_1$ and $s_2$ of Figure~\ref{fig:gap} gives 
\begin{equation}
\begin{aligned}
\label{eq:dist_fitness_10}
&F(g) = \\
&f \bigg( \sum_{n=0}^{10} B(n; 10, 0.15) \cdot \\ 
&\frac{1}{\binom{10}{n}} \sum_{e \in E_n:w_{6,10}(e)<3} \gamma \bigg( H \big( G(g, e \odot s_1),s_1  \big)\\
&+H \big( G(g, e \odot s_2),s_2  \big) \bigg) \bigg)
\end{aligned}
\end{equation}
where $w_{6,10}(e)$ denotes the weight of (number of -1's in) the second half of $e$. 

Substituting values 1 and $0.03125$ into Equation~\ref{eq:dist_fitness_10} and rearranging gives
\begin{equation}
\begin{aligned}
\label{eq:dist_fitness_10_spec}
&F(g) = f \bigg( \sum_{n=0}^{10} B(n; 10, 0.15) \cdot \\ 
&\binom{10}{n} \cdot \bigg( \sum_{e:w_{6,10}(e)<3 } \cdot 1 + \sum_{e:w_{6,10}(e)>2 } \cdot 0.03125 \bigg) \bigg) \\  
\end{aligned}
\end{equation}

Table \ref{tb:prob_not_back} summarises how many elementary perturbations of each total weight have second half weights $< 3$ (resp. $>4$).
Substituting its values into Equation~\ref{eq:dist_fitness_10_spec} gives a fitness bound to four decimal places of $0.9462$. 

\subsection{Comparing Runs}
\label{subsubsec:Runcompare}

The treatments examined in this paper use a mixture of stochastic and distributional evaluation. How can we fairly compare them? In the long run, if the same individual is repeatedly re-evaluated using stochastic evaluation, the mean fitness must converge to the distributional fitness; but in any particular case the stochastically evaluated fitness may be above or below the distributional fitness of the same individual. This would just induce noise in any comparisons (itself undesirable). However there is a further complication. We typically wish to compare the end-of-run best fitness achieved, which introduces a bias: even if the same individual is produced as that best individual: since it was chosen as the best individual in the generation, in a stochastic evaluation run its fitness is more likely to have been stochastically evaluated in the upper part of the fitness distribution. To eliminate this problem, for all comparisons, in all tables, and in all figures, except where explicitly mentioned, we always present the distributionally-evaluated fitness for an individual, even if it is evolutionarily evaluated using stochastic evaluation. This has the effect that for stochastic evaluation runs, figures showing the course of evoluation are not showing the actual fitness that was used in the evolution, but rather its distributional equivalent. The effect on a specific run could potentially be substantial.  However our results are typically averaged over 100 runs for a treatment. In this case, the stochastic variations largely cancel out. 

\subsection{Fitness Structure}
\label{subsubsec:FitStruct}

\begin{figure}
	\centering
	\includegraphics[width=\linewidth]{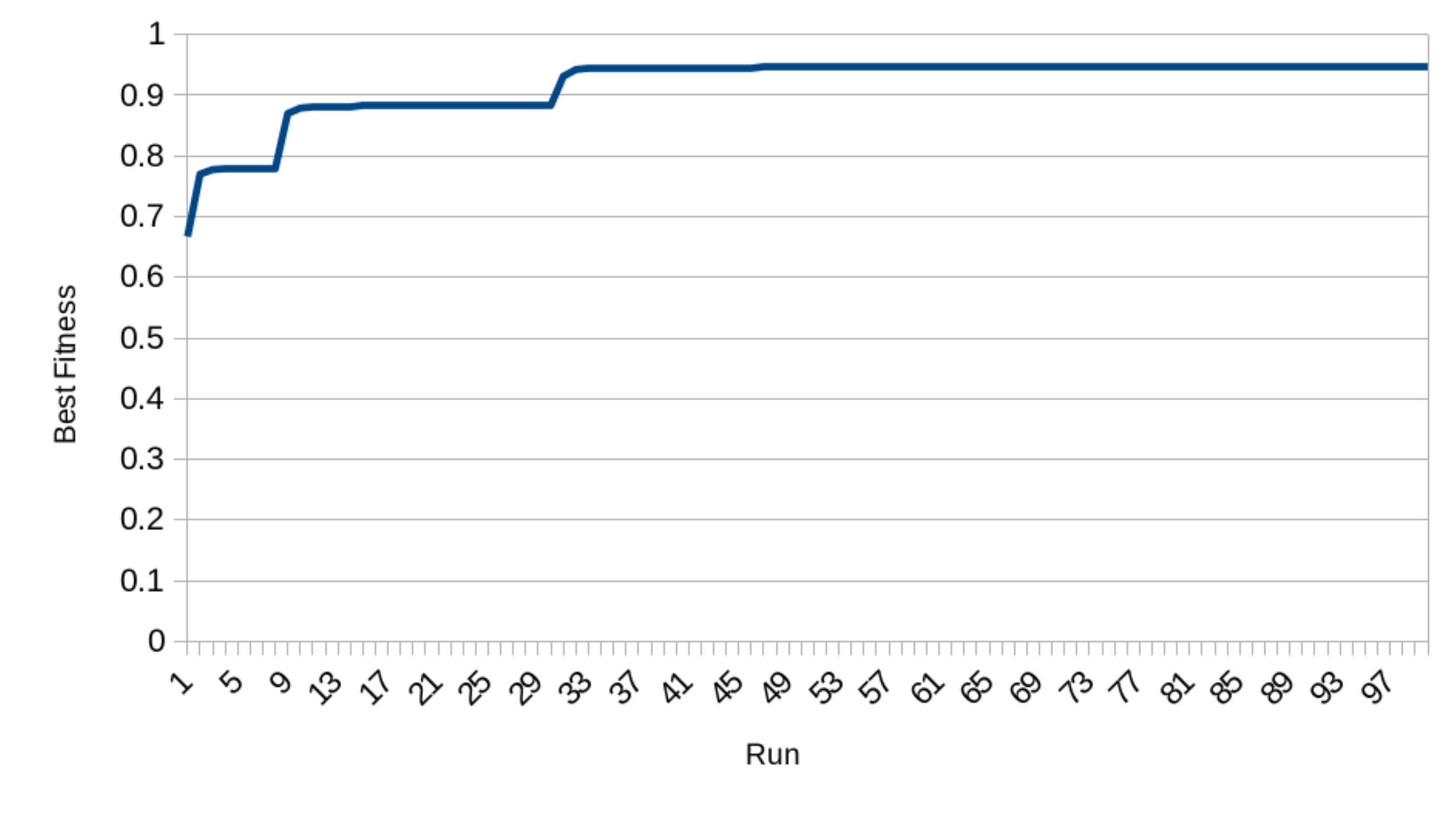}
	\caption{Typical End-of-Run Best Fitness Distribution over 100 Runs}
	\label{fig:OrderedHisto}
\end{figure}
Figure \ref{fig:OrderedHisto} shows a typical ordered histogram over 100 runs of the mean best distributional fitness achieved in the final generation (in this case, using distributional fitness evaluation as the objective). It is worth noting that the fitnesses are strongly suggestive of a plateau landscape (and in more detail, of multiple nested plateaux with further fine structure imposed on top). Essentially, this reflects the structure of the fitness evaluation: the primary plateau structure is determined by the success rate in restoring perturbations of size 1, there is a finer structure for perturbations of size 2, and so on for sizes 3 and 4. 

The consequence of this structure is that the primary metric we use in this paper, the mean over 100 runs of the  best fitness in a generation, and in particular the final generation, is primarily determined by performance on the size 1 perturbations. This should be borne in mind in viewing the figures. We note also that if a run achieves optimum fitness on the 10 size 1 perturbations, it usually achieves optimal fitness on all perturbations. This strongly suggests that, at least close to global optima, the plateaux of different scales (i.e. corresponding to different-sized perturbations) are not in conflict: the problems are not deceptive in this case. However we see more fine structure when a run only succeeds with a smaller number of size 1 perturbations, suggesting that there may be more conflict between layers (i.e. greater deception) further from the global optima.

\section{Distributional GRN Model}
\label{sec:dist_alg}
In this section, we present D-GRN, a distributional framework for A. Wagner's GRN model under distributional fitness evaluation, and briefly delineate the mutation and recombination operators. 

\subsection{Algorithmic Framework of D-GRN} 
  \scalebox{0.55}{
    \begin{minipage}{0.5\linewidth}
\begin{algorithm}[H]

	\caption{Algorithm Framework of D-GRN}
	\label{alg:das_grn}
	\KwIn{algorithm hyperparameters}
	\DontPrintSemicolon
	population size $\Pi$ of  size\;
	mutation rate M of rate (capital $\mu$)\;
	crossover rate X of rate (capital $\chi$)\;
	generations $\Gamma$ of number\;
	per-location perturbation rate E of rate (capital $\epsilon$)\;
	evolutionary tournament size T of  size (capital $\tau$)\;

	\KwOut{robust networks}
	
	\PrintSemicolon
	\tcp{Initialisation}
	crossover size $N_{\chi} = {X.\Pi} / 2$\;
	generate $\Pi$ feasible networks randomly\;
	save generated individuals in the population $P$\;
	get distributional fitness of each individual $I_i \in P$ \;

	\tcp{Loop until terminating condition}
	\For{$i=1$ to $\Gamma$} {
		Reset $P'=\Phi$\;	

		\tcp{Crossover}
		\For{$j=1$ to $N_{\chi}$}{
			\For{$l=1$ to 2}{
				\For{$k=1$ to T}{
					uniformly sample individual $I_{j,l,k} \sim P$\;
				}
				select $I_{j,l} = \text{argmax}_{k \in \{1,\ldots,\text{T}\}} F(I_{j,l,k})$\;
			}
			generate $I_{j,3}$ and $I_{j,4}$ from $I_{j,1}$ and $I_{j,2}$ by diagonal recombination\;
			$P' = P' \bigcup \lbrace I_{j,3}, I_{j,4}\rbrace$\;
		}
		\tcp{Copying}
		\For{$j=2N_{\chi} + 1$ to $\Pi$}{
			uniformly sample individual $I_j \sim P$\;
			$P' = P' \bigcup \lbrace I_{j}\rbrace$\;
		}
		\tcp{Mutation}
		\ForEach{individual $I_i$ in $P'$}{
			uniformly sample real $r \sim [0,1)$\;
			\If{$r<M$}{
				biasedly mutate $I_i$\;
			}
		}
		
		\tcp{Updating}
		get distributional fitness of each individual $I_i \in P'$\;
		update $P = P'$ \;
	}
\end{algorithm}
\end{minipage}
}  
\newpage

Algorithm \ref{alg:das_grn} presents a high-level structure for a typical haploid genetic algorithm; in this work, the instantiations of the recombination and mutation operators are slightly atypical, so we detail them here.

\subsection{Recombination}
\label{subsec:recomb}
\begin{figure}[htb]
	\centering
	\begin{subfigure}[b]{0.48\linewidth}
		\includegraphics[width=\linewidth]{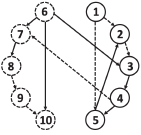}
		\caption{Parental GRN 1}
	\end{subfigure}
	\begin{subfigure}[b]{0.48\linewidth}
		\includegraphics[width=\linewidth]{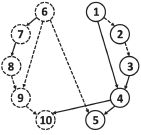}
		\caption{Parental GRN 2}
	\end{subfigure}
	\begin{subfigure}[b]{0.48\linewidth}
		\includegraphics[width=\linewidth]{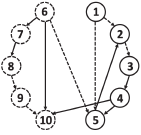}
		\caption{Offspring GRN 1}
	\end{subfigure}
	\begin{subfigure}[b]{0.48\linewidth}
		\includegraphics[width=\linewidth]{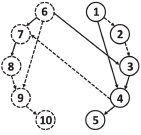}
		\caption{Offspring GRN 2}
	\end{subfigure}
	\caption{Illustration of the diagonal recombination, with the pivot index as $5$. Solid and dashed circles represent different modules. Solid and dashed arrows stand for activation and repression respectively. }
\label{fig:diagonal_crossover}
\end{figure}
In~\citep{10.1162/isal_a_00166}, we introduced and experimentally investigated diagonal recombination. 
Given two parental GRNs $A_1[1, \ldots,  10]$ and $A_2[1, \ldots,  10]$, diagonal recombination proceeds by first sampling a pivot point $i$ from $\lbrace 1, \ldots, 10 \rbrace$, then preserving the two sub-matrices
$A_1[1 \ldots i-1, 1 \ldots i-1]$ and $A_1[i \ldots 10, i \ldots 10]$,
while exchanging the remainder of corresponding locations between $A_1$ and $A_2$. This is illustrated in figure~\ref{fig:diagonal_crossover}. We note that if $i$ is sampled as $1$, the corresponding recombination is a null operation; and that recombination does not change the total number of interactions, and therefore the mean, in the population.

\subsection{Mutation}
\label{subsec:mutation}
The mutation operator of~\citet{espinosa2010specialization} biases the edge density to a relatively low level.
A node in the GRN has a probability $\mu$ to mutate every generation; if it mutates, it can either lose or gain an interaction. 

With the mutation rate setting of~\citet{espinosa2010specialization} of $\mu=0.05$, the probability that an individual is unchanged by mutation is drawn from the binomial B(10,0.05), having a probability of $p \approx 0.5$. This is relatively high for an evolutionary algorithm, substantially increasing the headline replication rate, and decreasing exploration. This may not much affect algorithms using the sampling evaluation of~\citet{espinosa2010specialization}, since the noisy evaluation provides an additional source of stochasticity, but is sub-optimal for an algorithm with deterministic evaluation, such as the distributional algorithm presented here. For that reason, we use a higher mutation rate of $\mu=0.2$ in these comparisons.

The probability for a node to lose an interaction (a nonzero value to change to zero) is defined as
\begin{equation}
\label{eq:edge_bias}
p(u)=\frac{4r_{u}}{4r_{u} + N - r_{u}}
\end{equation}
where $N$ is the number of genes in a gene activation pattern of a target, and $r_{u}$ is the number of regulators of gene $u$~\citep{espinosa2010specialization}, i.e. nonzero values in column $u$. Conversely, the probability for a gene $u$ to gain an interaction (i.e. for a zero value in row $u$ to become nonzero) is $1-p(u)$. The neutral point of this bias can be computed as: 
\begin{equation*}
\begin{aligned}
p(u) = 1 - p(u) \Rightarrow r_u = \frac{N}{5}
\end{aligned}
\end{equation*}
The bias acts to maintain the sparsity of the network at around this value, which research in computational biology suggests is essential to induce modularity~\citep{leclerc2008survival}.

\clearpage 
\begin{table}[!htbp]
	\centering
	\resizebox{0.8\textwidth}{!}{
	\begin{threeparttable}
	\caption{Explanations of simulation parameters}
	\label{table:parmexplain}
	\begin{tabular}{ll}
		\toprule
		Target Pattern & pattern to be perturbed then recovered \\
		Introduction Stage & generation target is introduced \\
		Pattern Size & $N$, \texttt{\#} of locations in activation pattern \\
		GRN Size & $N \times N$, size of a GRN (genotype) \\
		Initial Density & initial density of edges in the GRN \\
		\texttt{\#} Perturbations & \texttt{\#} of perturbations of each target\tnote{a}\\
		Perturbation Rate & expected proportion of corrupted genes \\
		Population Size & \texttt{\#} of individuals in the population\\
		Mutation Rate & probability GRN node gains/loses  an interaction\tnote{b} \\
		Activation Rate & proportion of new interactions that are activations\tnote{c} \\
		Crossover Rate & proportion of individuals crossed over\tnote{d} \\
		Crossover Type & type of crossover (recombination) used\\
		Selection (size) & type (and size) of selection \\
		Replication Rate & proportion generation simply copied \\
		Max. generation & generation to terminate simulation   \\
		Trials per Treatment & \texttt{\#} of trials for comparing treatments \\
		Significance Test & statistical test for comparing treatments \\ 
		\bottomrule    
	\end{tabular}
	\begin{tablenotes}
	\vspace{0.5 em}
            \item{a} For distributional evaluation, this will be $2^N$.\\
            \item{b} For compatibility with the terminology of~\citet{espinosa2010specialization}. In EC terms:\\
            	\hspace*{2 em} the per-gene mutation rate in the evolving GRN is $1/N$ of this\\
	    	\hspace*{2 em} the per-individual mutation rate is $N$ times this\\
	    \item{c} Gained interactions are either activations or repressions.\\
	    \item{d} Some crossovers may be ineffective, as previously discussed
        \end{tablenotes}
	\end{threeparttable}
	}
\end{table}

\clearpage 

\begin{figure}[h]
	\centering
	\begin{subfigure}[b]{0.6\linewidth}
		\includegraphics[width=\linewidth]{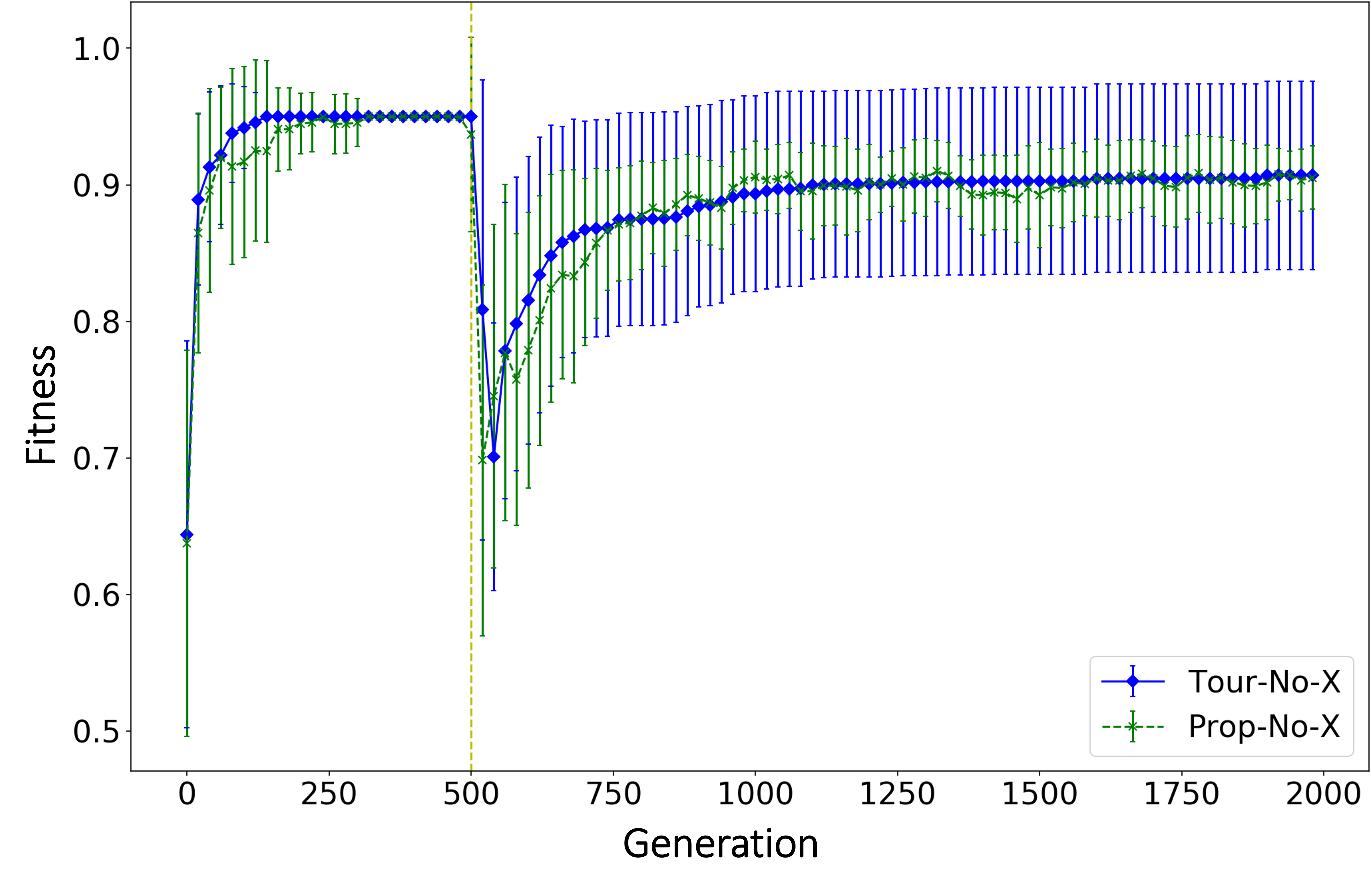}
		\caption{Best Fitness in Every Generation.}
	\end{subfigure}
	\begin{subfigure}[b]{0.6\linewidth}
		\includegraphics[width=\linewidth]{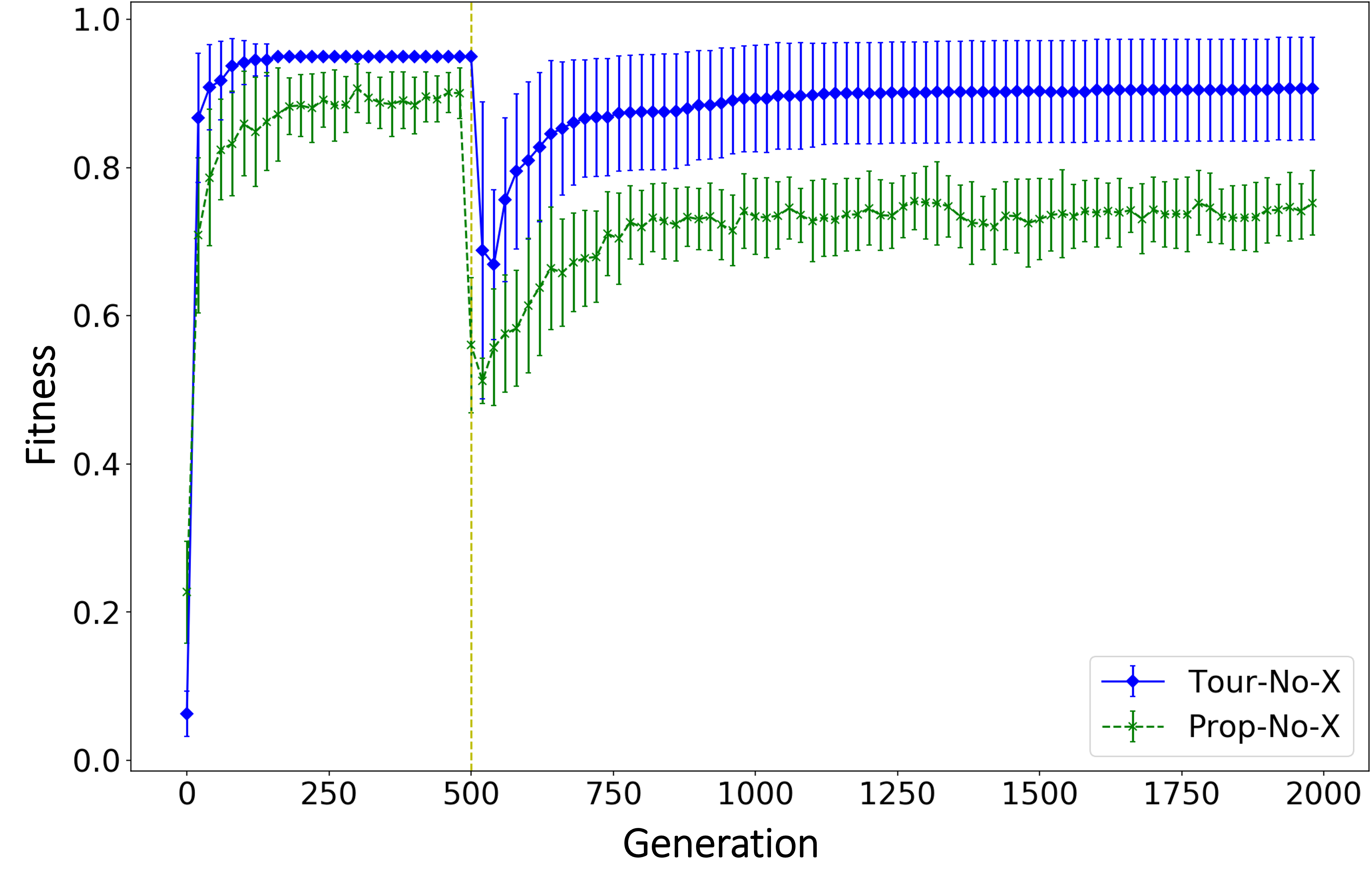}
		\caption{Median Fitness in Every Generation.}
	\end{subfigure}
	\caption{Best and median fitness over generations of evolution (without recombination) comparing the performance of tournament and proportional selections. Blue dots stand for tournament and green crosses proportional. }
	\label{fig:tour_prop_fit_comp_no_x}
\end{figure}

\begin{figure}[htbp]
	\centering
	\begin{subfigure}[b]{0.48\linewidth}
		\includegraphics[width=\linewidth]{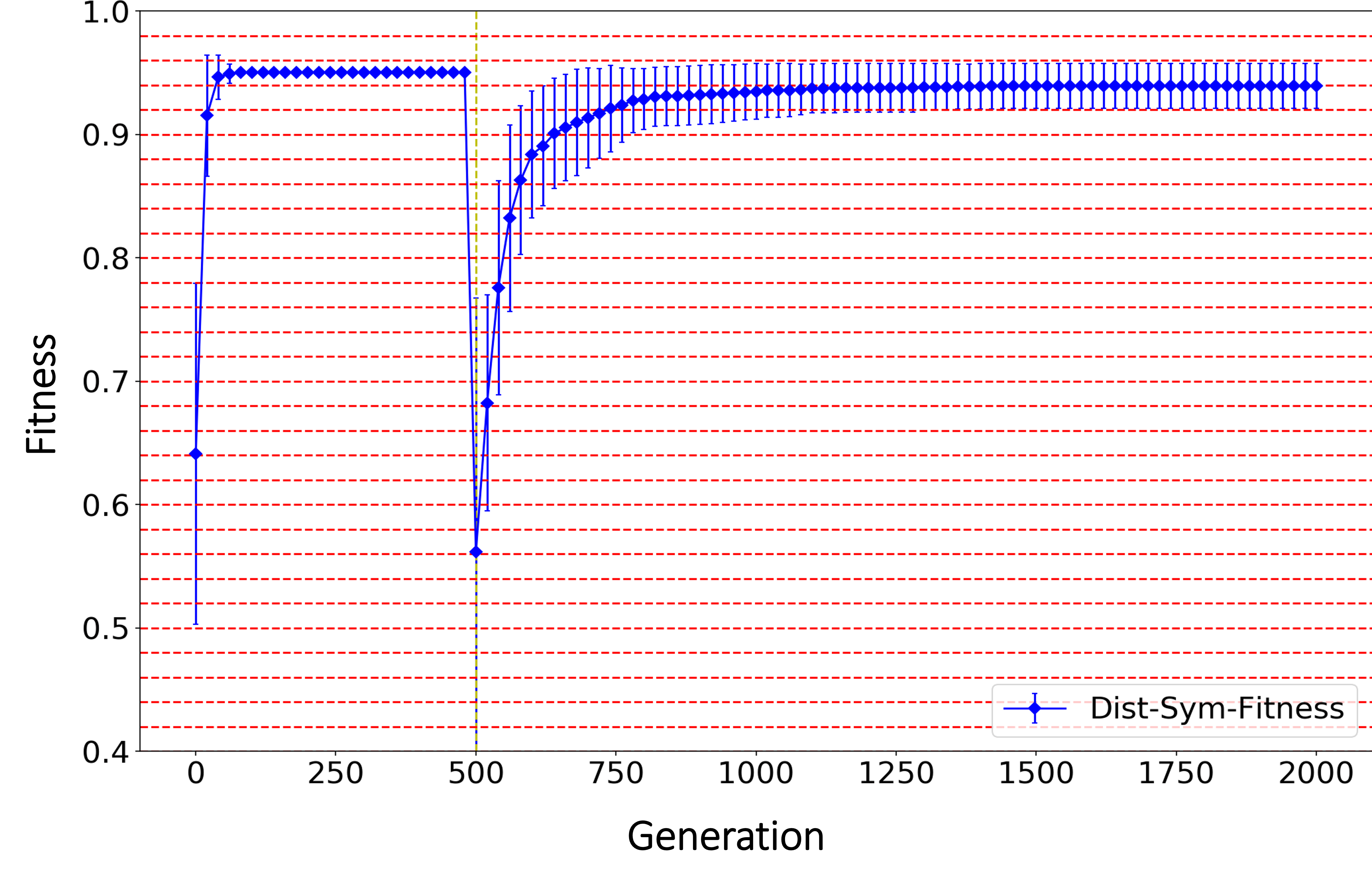}
		\caption{Distributional Fitness}
	\end{subfigure}
	\begin{subfigure}[b]{0.48\linewidth}
		\includegraphics[width=\linewidth]{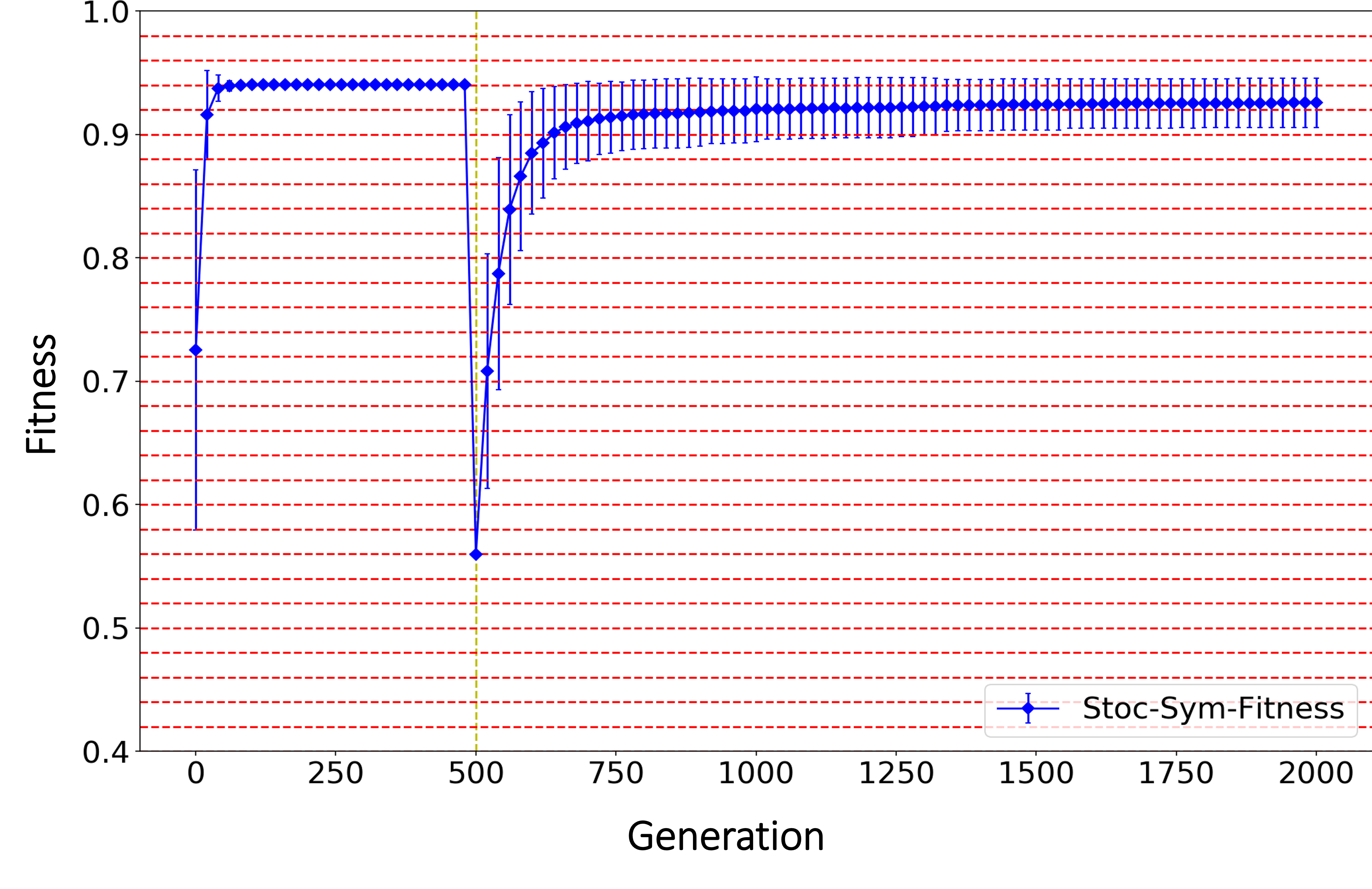}
		\caption{Stochastic Fitness}
	\end{subfigure}
	\begin{subfigure}[b]{0.48\linewidth}
		\includegraphics[width=\linewidth]{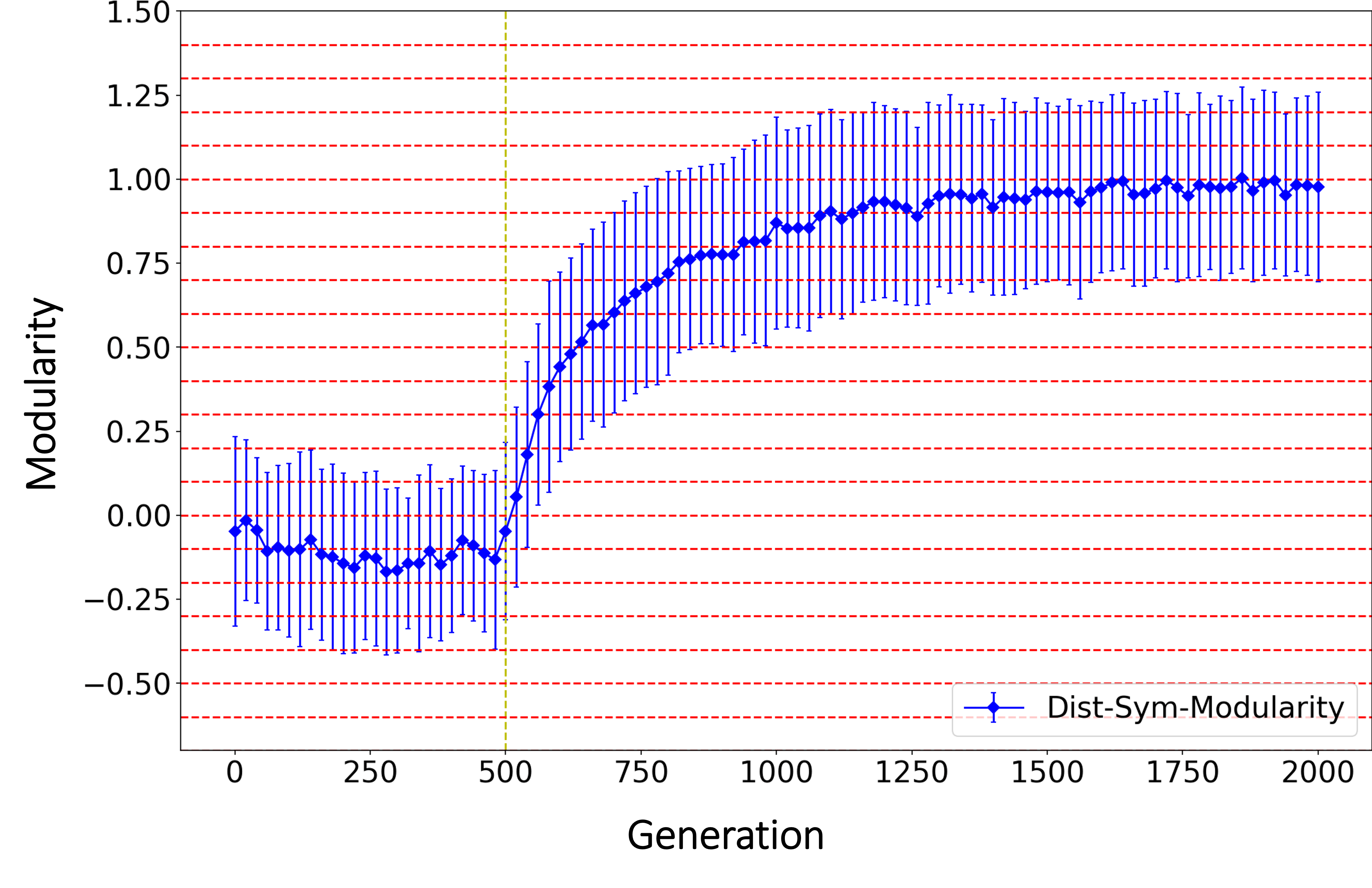}
		\caption{Distributional Modularity}
	\end{subfigure}
	\begin{subfigure}[b]{0.48\linewidth}
		\includegraphics[width=\linewidth]{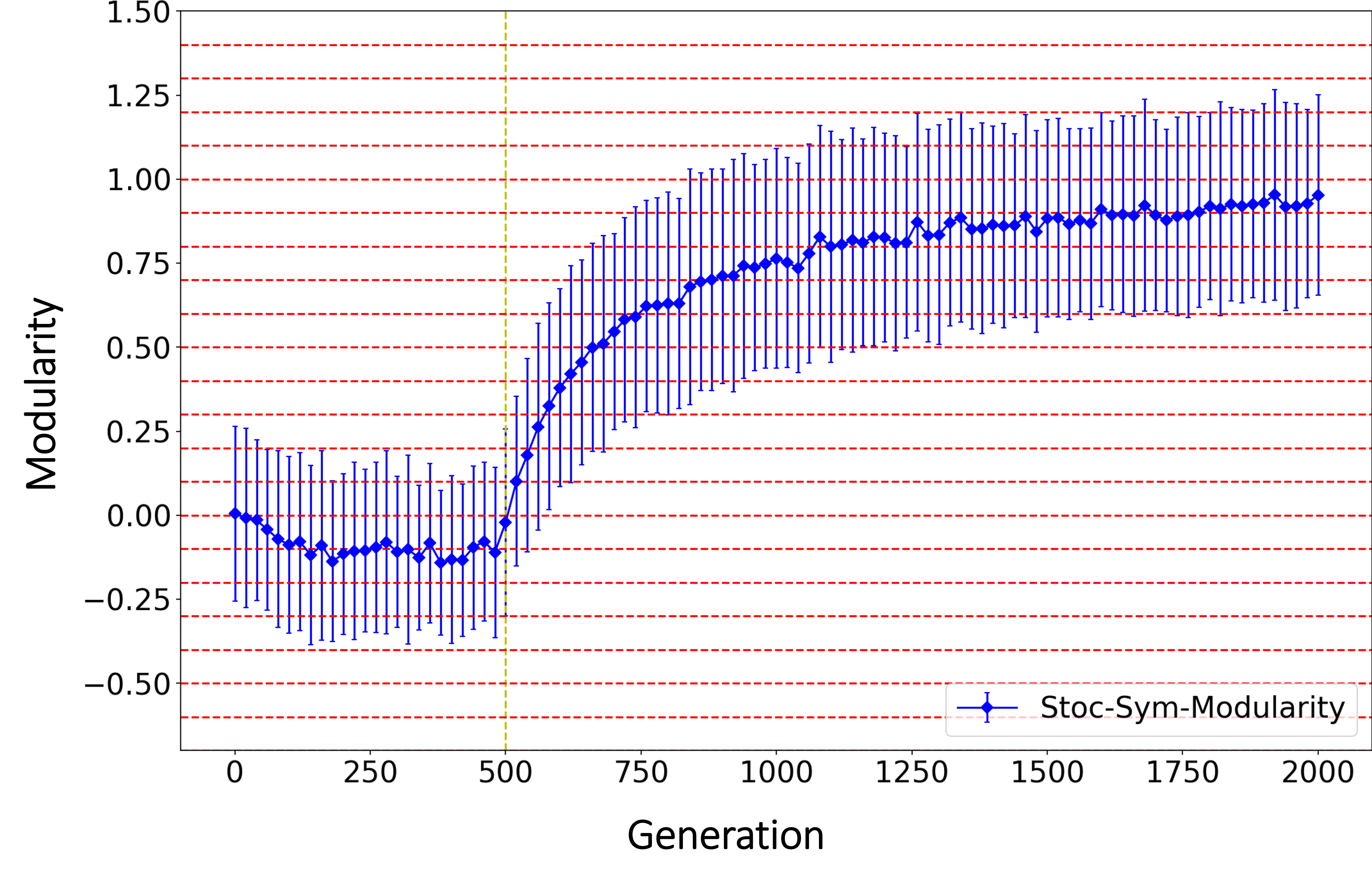}
		\caption{Stochastic Modularity}
	\end{subfigure}
	\caption{Evolutionary Progress for each Generation, Means over 100 Trials.
Left to Right: Distributional vs Stochastic Evaluation. 
Top: Fitness; Bottom: Modularity.
Vertical bars represent one standard deviation. The vertical dashed line at generation 500 indicates addition of the second activation pattern.}
	\label{fig:dist_stoc_fit}
\end{figure}

\begin{figure}[h]
	\centering
	\begin{subfigure}[b]{0.48\linewidth}
		\includegraphics[width=\linewidth]{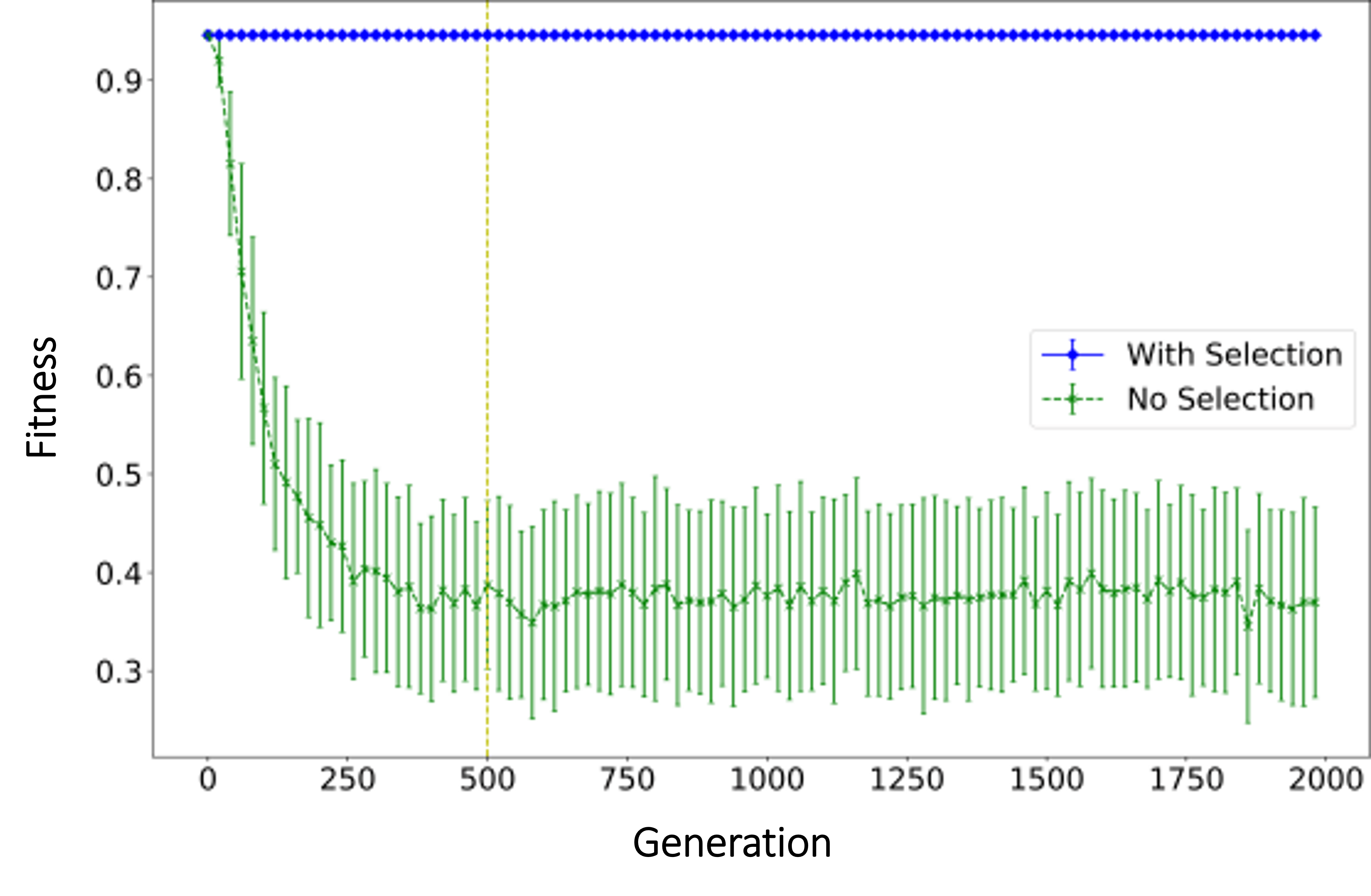}
		\caption{Fitness}
	\end{subfigure}
	\begin{subfigure}[b]{0.48\linewidth}
		\includegraphics[width=\linewidth]{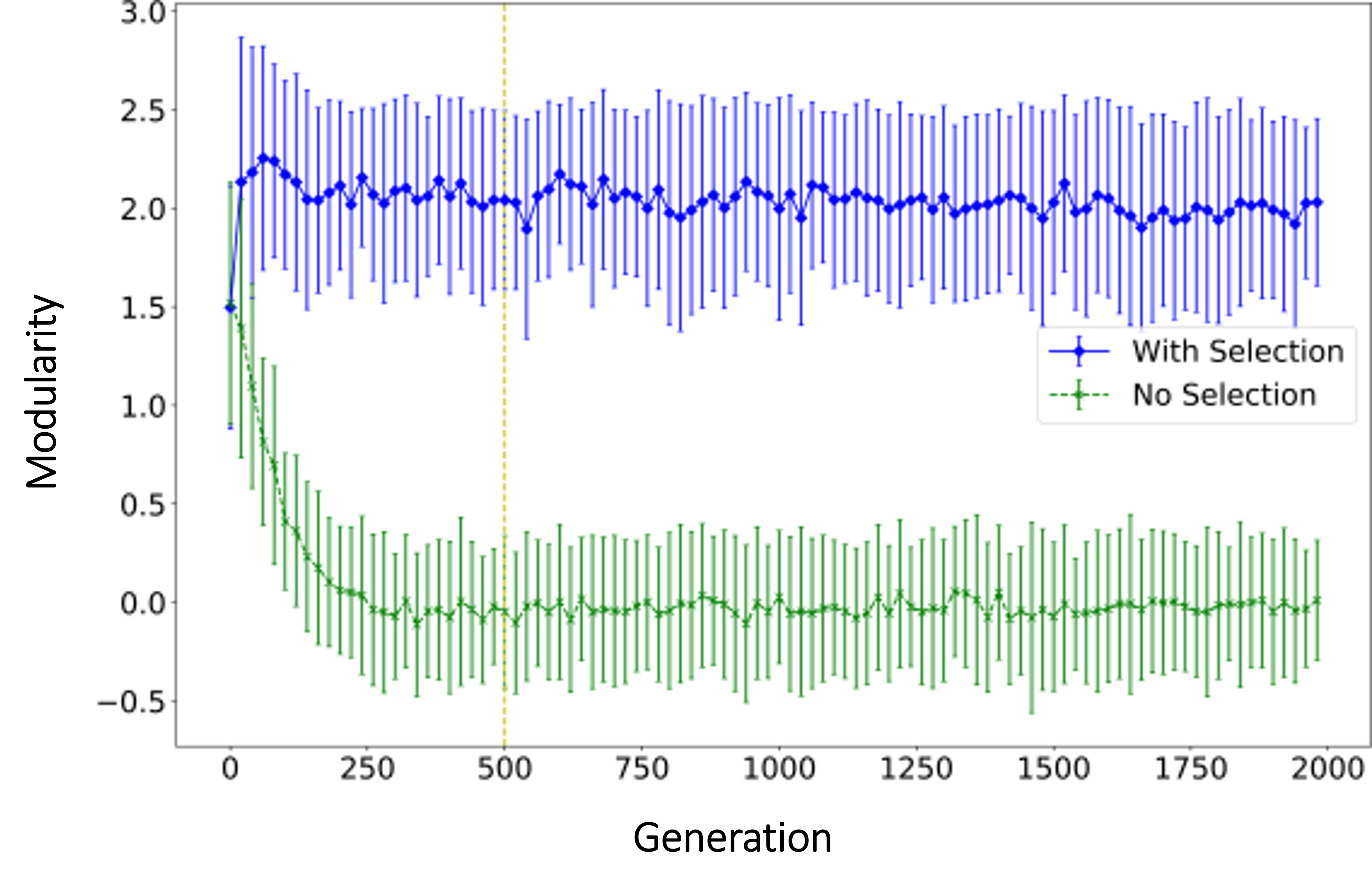}
		\caption{Modularity}
	\end{subfigure}
	\caption{Progress of Evolution of an Initially Optimal Population under Two-Target Selection vs No Selection:\\ Fitness and Modularity}
	\label{fig:avg__fit_mod_perfect_module_1}
\end{figure}

\begin{figure}[h]
	\centering
	\includegraphics[width=0.7\linewidth]{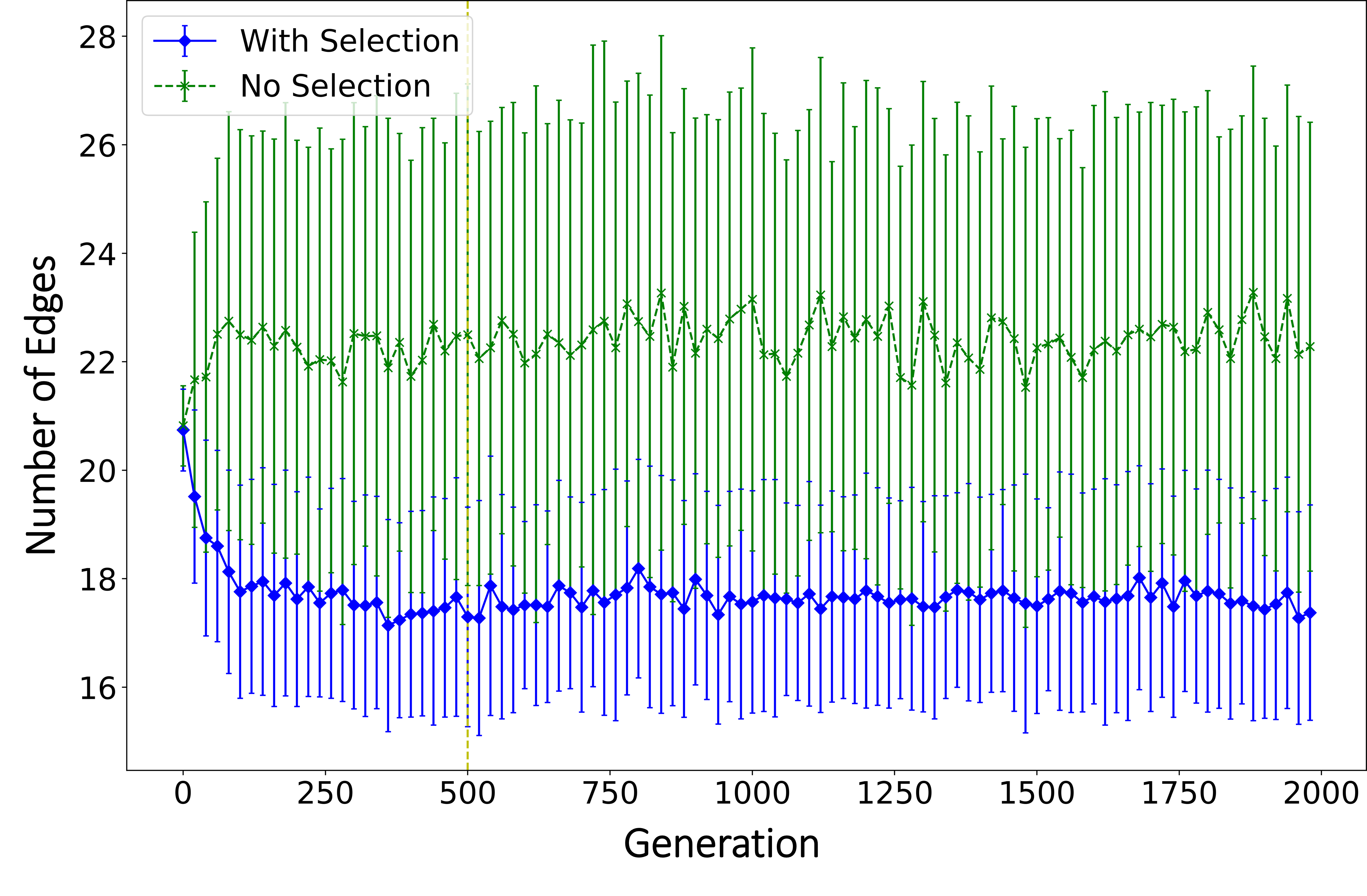}
	\caption{Progress of Evolution of an Initially Optimal Population under Two-Target Selection vs No Selection:\\ Number of Edges}
	\label{fig:avg_edge_perfect_module_1}
\end{figure}

\section{Experimental Setup}
\label{sec:exp}
All simulation code was implemented in Java 1.8.0 and Python 2.7.10. All programs are publicly available at \url{https://github.com/ZhenyueQin/Project-Maotai-Modularity}. 
We use the Mann-Whitney significance test in all comparisons. 

\subsection{Parameter Tables and Explanations}
The specific targets $T$ used in the experiments appear in Table~\ref{table:genactiv}. Evolutionary and other simulation parameters for the main body of experiments are specified in Table~\ref{table:simparams} and explained in Table~\ref{table:parmexplain}. A few variations on these will be specified in the relevant context.

\subsection{Population Initialisation}
A population in our simulated evolution consists of 100 individuals, each being a GRN as defined in Eq \ref{eq:grn}. During initialisation, each individual in the population will randomly generate 20 edges with arbitrary directions. The chosen number 20 comes from Eq. \ref{eq:edge_bias} that biases towards sparse networks. Specifically, Eq. \ref{eq:edge_bias} biases toward each of the $10$ genes having $2$ regulators, giving an overall bias toward $20$ active regulators per GRN. However the number of possible edges ranges from $0$ to $100$ (the edges being directional), so the probability mass function for number of edges, despite having a peak at $p(e=20)$, will skew to the right. That is, 
\begin{equation*}
\begin{aligned}
\sum_{e=0}^{19}p(e) < \sum_{e=21}^{100}p(e)
\end{aligned}
\end{equation*}
Thus even without selection, the number of edges will be expected to increase slightly during a run. Running such experiments, we found that the average GRN edge number converged to approximately $22$. 

\subsection{Selection Scheme: Tournament vs Proportional Selections}
In evolutionary computation, there are two prevalent selection schemes: tournament and proportional. Although~\citet{espinosa2010specialization} uses proportional, we use tournament in this paper because it gives similar results to proportional selection, but is better suited to incremental evaluation, a strategy we plan to use in extending this work to larger targets. We ran experiments comparing tournament and proportional selection. The results are in \autoref{fig:tour_prop_fit_comp_no_x}. We found tournament selection generated higher median fitness over the generations, and also better preserved fitness (\ie fluctuations were reduced). To ensure a fair comparison, in this experiment we turned crossover off as in ~\citep{espinosa2010specialization}.

\subsection{Modularity Metric}
\label{subsec:mod_metric}
We adopt the normalised Q scoring system to quantify modularity in a GRN, ultimately based on the definition proposed by ~\citet{newman2004finding}. Briefly, this quantity is defined as the difference between, on the one hand, the ratio of the number of edges in the network connecting nodes within modules to the total number of edges, and on the other, the expected value for the same quantity for a randomly connected network with the same edge density~\citep{kashtan2005spontaneous}. Formally, $Q$ is calculated as 
\begin{equation}
\label{eq:modularity_q}
Q = \sum_{i=1}^{K}[\frac{l_i}{L} - (\frac{d_i}{2L})^2]
\end{equation}
where $i$ represents one of the $K$ potential modules within a network, $L$ is the total number of connections in a network, $l_i$ stands for the number of interactions in the module $i$, and $d_i$ is the sum of degrees of all th$   $e nodes in module $i$~\citep{espinosa2010specialization}. The value $Q$ will lie in the range of $[-0.5, 0.5]$ for our module structure.

Unfortunately the expected value of Q varies according to the density of edges. In order to eliminate the effects of variations in total edge numbers within a GRN, and provide fair comparisons, we normalise Q as in Equation~\ref{eq:norm_q}, following the spirit of \citet{kashtan2005spontaneous}.
\begin{equation}
\label{eq:norm_q}
Q_n = \frac{Q - Q_{ran}}{Q_{max} - Q_{ran}}
\end{equation}
where $Q$ is the modularity Q value obtained from Equation~\ref{eq:modularity_q} for the specific network $\{ V, E \}$, $Q_{ran}$ is the average $Q$ value of 10,000 random networks with the same number of vertices $V$ and edges $E$ as the network $\{ V, E \}$, and $Q_{max}$ stands for the maximum $Q$ value in these 10,000 random networks. This normalised $Q_n$ shows us how modular our network is by comparing it to sampled highly modular and random networks with the same attributes~\citep{espinosa2010specialization}. 

\section{Experimental Results and Analysis: Stochastic vs Deterministic Evaluation}
\label{sec:exp_res}
In this set of experiments, we compare the behaviour of the algorithm using stochastic evaluation with that using distributional evaluation. From the perspective of the particular domain, this helps to understand to what extent aspects of the behaviour (for example, emergence or non-emergence of modularity) are a consequence of the fitness landscape of the problem, and to what extent they derive from the noise effects of random sampling imposed upon that landscape. From an evolutionary biology perspective, they allow us to compare the behaviour of small population and perturbation sample sizes (computationally tractable but generally biologically implausible) against the smoothing effect of effectively infinite perturbation samples. Because a key effect of increasing either population size or perturbation sample size is to smooth behaviour, this can help to gain some insight into what we might expect from more biologically realistic (but computationally infeasible) population sizes. Finally, from a methodological perspective, this section illustrates what is feasible if the distribution underlying the noise is small enough to be directly computable (in this case, the GRN target is sufficiently short). Without further theoretical advances, distributional evaluation would not be computationally tractable if the target, and thus the distribution, was much larger.

We conducted 100 independent evolutionary simulations for distributional and for stochastic fitness evaluation. We collected the fittest GRNs in each generation and evaluated their fitnesses and modularities. We remind readers that these results are reported using the distributional fitness, even for runs using stochastic evaluation. An important consequence is that we know, even for stochastic runs, whether we have actually found a true optimum.

\subsection{Effects of Stochastic Evaluation on Evolutionary Efficiency}
\subsubsection{Results}

Table \ref{table:dist_stoc_fit} 
shows the outcomes for fitness and modularity in the final generation, 
while figure~\ref{fig:dist_stoc_fit} shows the evolution of fitness and modularity over the generations. 

Distributional fitness evaluation leads to a small, but statistically highly significant, increase in final-generation fitness when compared with stochastic evaluation. A large part of the reason is that the stochastic algorithm tends to overestimate, and thus retain, anomalously high sampled values. Thus, at least with these evolutionary settings, the dynamicity imposed by stochastic evaluation has a deleterious effect. From a different perspective, Distributional evaluation achieves a (distributional) fitness only 0.0067 below the theoretical bound of 0.9462, while distributional fitness evaluation is 0.0206 below. Thus while the fitness difference is small, the difference from the attainable optimum is a factor of three larger.  For completeness, in the table we also show the fitness that the stochastic best individuals recorded using stochastic evaluation (in fact the difference is small). The fitness effect is mirrored by a small, but non-significant, improvement in modularity.

Figure~\ref{fig:dist_stoc_fit} shows that fitness and modularity evolve very similarly for the two evaluation methods up to the change of target. Subsequently, the fitness graphs remain similar (although stochastic fitness is always slightly below distributional), but the modularity from stochastic evaluation increases more slowly after the target change than the distributional. It is tempting to speculate that the latter might result from reduced search eagerness due to the stochasticity of the fitness function, but if that is the reason, it is surprising that the effect on fitness is not more pronounced.

\subsection{Deterministic Fitness Evaluation with Complete Sampling Can Help Better Analyse GRN Edge Functions}
In previous work~\citep{qin2018don, 10.1162/isal_a_00166}, 
we reported that the simple procedure of manually removing inter-module edges from evolved highly fit but non-modular solutions could, in the
 majority of cases (24/40), further improve their fitnesses. We were puzzled by this phenomenon, because we expected that this procedure -- favoured by the mutation bias -- should be easily followed by an evolutionary algorithm. Using the deterministic distributional fitness evaluation, we determined that this scenario was mainly due to the stochasticity of the original fitness evaluation. Over 100 runs with distributional fitness evaluation, we collected the 100 fittest GRNs from the final generation. Among these, we manually removed any inter-module edges and measured the resulting fitnesses. Under distributional evaluation, only 7/100 had an improved fitness. However under stochastic evaluation, the imposed noise meant that 36/100 GRNs appeared to improve their fitness by undergoing this procedure. Thus in most cases, the improvement was purely illusory; the few cases where there was a real (i.e. distributional) improvement in fitness were more than counterbalanced by the large number of cases where there was a deterioration.

\section{Further Characteristics of the GRN Model}
We regard the results of the previous section as the primary result of this work, disentangling the roles of the underlying fitness landscape of~\citet{espinosa2010specialization}, and of their particular (sampling based) method for evaluating it. In the course of this work, however, we also determined further characteristics of the problem domain, which we present here for completeness.

As previously noted, one important advantage of the distributional perspective is the ability to determine when we have actually attained a global optimum. This then permits a kind of reverse research path: instead of examining how an algorithm searches for the optimum, we can instead examine characteristics of the optima it finds, and also what can happen once the algorithm has attained an optimum. 

We note that the settings used in this section were as described earlier, with the exception that for compatibility with~\citep{qin2018don}, we used mutation and crossover rates of 0.2 and 1.0 rather than the more typical rates of 0.2 and 0.2 shown in Table~\ref{table:simparams}.

\subsection{Stochastic Search Reliably Maintains Global Optima}

Since we had noted that search based on stochastic sampling-based search had difficulty in precisely locating highly fit, modular global optima even when it was sampling very close to their vicinity, we wondered whether perhaps the evaluation stochasticity might prevent it maintaining global optima when it had found them. So we essentially ran the search backward, by starting with a population of global optima.
Instead of randomly generating initial GRNs at the starting point of the evolution, we manually initialised a population of different GRNs found in previous experiments, with two characteristics: 
\begin{enumerate}
	\item the fitness of each is globally optimal. 
	\item Its modularity is perfect, i.e., there are no edges between modules.
\end{enumerate}

We eliminated Phase \rom{1} of the evolution, using only Phase \rom{2}, \ie the GRNs had to simultaneously regulate the two activation patterns of Figure~\ref{table:genactiv}. As a comparator, we used an algorithm with no selection. As Figure~\ref{fig:avg__fit_mod_perfect_module_1} indicates, the system was able to preserve optimal fitness, and also maintained modularity at a relatively high level.

Intriguingly, the number of edges actually fell substantially, (Figure~\ref{fig:avg_edge_perfect_module_1}), suggesting an intrinsic parsimony that did not arise from, and in fact acted in the opposite direction to, the bias imparted by the size-biased mutation operator. Compared with the selection-free treatment, the size was significantly and very substantially smaller ($17.8046$ vs $23.3545$, Mann Whitney Test: $p=1.3123 \times 10^{-17}$). Since the individuals measured are already almost completely modular, the edges lost must be intra- rather than inter-modular edges. Plausible hypotheses suggest that either mutation or diagonal crossove (perhaps both) may be more disruptive to the fitness of higher-density GRNs, and thus may indirectly provide a parsimony pressure.

\subsection{Underlying Patterns of Optimal Modules}
\begin{equation}
\label{eq:patterned_2_grn_1}
\begin{bmatrix}
+1 & -1 & +1 & -1 & +1 \\
-1 & +1 & -1 & +1 & -1 \\ 
+1 & -1 & +1 & -1 & +1 \\
-1 & +1 & -1 & +1 & -1 \\
+1 & -1 & +1 & -1 & +1 \\
\end{bmatrix}
\end{equation}
Let us consider an optimal, maximally modular GRN. It is a $10 \times 10$ matrix of four quadrants; being modular, the top left and bottom right modules must be empty, and the most interesting quadrant is the bottom right one, which (being optimal) must regulate the recovery of a perturbation to the Manhattan-closest of the two target patterns. Note that this can be viewed as a kind of majority-voting problem: if the perturbed pattern has three or more bits in common with a particular target, it should move to that target; otherwise it must have three or more bits in common with the alternate target, and should move there.

In examining fitness- and modularity-optimal examples of this quadrant, we noticed a pattern, which can be explained in terms of matrix~\ref{eq:patterned_2_grn_1}. First, we may note that this matrix is in fact optimal: if it is inserted as the bottom right quadrant of a modular GRN (i.e. one in which the top left and bottom right quadrants are empty), it perfectly regulates the second half of the target as required by the fitness function. But that is not all. 

Define a \emph{shadow} of a matrix as one in which some nonzero values are replaced by values which are closer (but in the same direction) to zero, and a discrete shadow as one in which the replaced values actually are zero. When we computed mean values across the matrices of optimal patterns that we found in series of runs, we generally found that the mean matrix had a bottom quadrant that was a shadow of matrix~\ref{eq:patterned_2_grn_1}: that is, solutions shadowed the structure of matrix~\ref{eq:patterned_2_grn_1}. Individual fully modular optimal solutions that we examined were generally Manhattan-close to discrete shadows; and if they did have entries in that bottom right quadrant that were counter to the pattern, they otherwise were 'denser' shadows of matrix~\ref{eq:patterned_2_grn_1} -- we conjecture that they countered any off-pattern entries with more of the on-pattern entries to restore the majority voting structure. There was also a tendency for non-modular solutions to have denser shadows of matrix~\ref{eq:patterned_2_grn_1}, which again we conjecture had the effect of countering non-modular influences with a greater influence from majority voting. Thus while there may be solutions (modular or non-modular) that do not follow the structure of matrix~\ref{eq:patterned_2_grn_1}, in general the solutions found by evolutionary algorithms do follow this structure. This has important implications for future research.

\section{Conclusions}
\citet{qin2018don} revealed a number of anomalous behaviours of a fairly typical genetic algorithm when it was used in the modularity emergence model of~\citet{espinosa2010specialization}. These consisted of:
\begin{enumerate}
	\item \label{item:elitism}Initially, failure to observe the emergence of modularity at all
	\item \label{item:mutrate}A simple sampling-based fixed fitness function did not lead to modularity
	\item \label{item:edgeremove}In many runs, taking the fittest solution found and applying a simple process of removing all intermodular edges resulted not only in complete modularity, but also higher fitness -- yet the algorithm was unable to find those solutions.
\end{enumerate}

\citet{qin2018don} had already traced the cause of anomaly~\ref{item:elitism} to too eager search due to the incorporation of elitism. This paper has resolved the remaining anomalies. Anomaly~\ref{item:mutrate} also stemmed from insufficient exploration, in this case due to a too-low mutation rate. Restoration of a more typical mutation rate restored the emergence of modularity. Anomaly~\ref{item:edgeremove} turned out, paradoxically, to be due to the very stochasticity that had originally hidden anomaly~\ref{item:mutrate}: noise in the fitness evaluation completely concealed from the algorithm the available fitness-increasing paths to modular solutions.

The resolution of these anomalies derived from the introduction of distribution-based evaluation, which enabled deeper analysis of the algorithm behaviour, in particular because it allowed reliable identification of the global optima of the problem. It also allowed us to determine that the emergence of modularity is purely a property of the fitness landscape defined by~\citet{espinosa2010specialization}; the stochastic noise resulting from sampling-based evaluation is not necessary for the emergence of modularity -- a conclusion that initially surprised us. We surmise that modularity will emerge under any suitable optimisation algorithm that can sufficiently approximate the fitness landscape while retaining sufficient exploratory power. 

We also revealed further important properties of the fitness landscape:
\begin{itemize}
	\item Failures of the algorithms to deliver optimal solutions are largely due to difficulty in locating them; it is not difficult for these algorithms to retain populations of highly fit, modular solutions once found
	\item The pattern of matrix~\ref{eq:patterned_2_grn_1} is a key component of most optimal solutions, with near Manhattan-neighbours to its shadows appearing in most optimal solutions, and in particular in modular ones
\end{itemize}

Overall, we view the introduction of a distributional perspective, and the specific application of distributional evaluation, as the most important contributions of this work. We believe that these approaches are applicable to a wide range of other sampling-based evolutionary simulations.

\subsection{Further Work}
The model of~\citet{espinosa2010specialization} was originally motivated, at least in part, by its extensibility to larger targets, and potential exploration of the emergence of modularity in more complex environments. Our own experience, and that of others~\citep{Adami2019perscom}, is disappointing: sampling methods seem  unable to generate modularity (or high fitness) with targets of length 20 or above. Yet such solutions do exist: we can readily extend solutions based on matrix~\ref{eq:patterned_2_grn_1} to targets whose size is any multiple of 5: they are of course completely modular; they are also highly fit (though the analysis of actual optimality becomes increasingly complex with target size, so they may not be global optima). So why does search not find them? We believe the answer to be the same as for the size 10 case. That is, naive sampling is too stochastic, the resulting noise drowning out the small variations in fitness in paths leading to good solutions. For targets of size 15, we could potentially instead apply our distributional fitness function, at the price of an approximately 32-fold increase in computational cost. Beyond that size, the computational complexity is simply overwhelming. 

Our hope is that future research will have more sophisticated sampling methods, using distribution-based estimation for smaller weight (polynomial-scaling) perturbations and only using sampling for the larger weight (exponential-scaling) perturbations that carry decreasing importance in the fitness estimation.
Unfortunately our efforts in this direction have not -- so far -- proven particularly successful~\citep{9308466}. In order to succeed, it may be necessary to reduce the perturbation rate (the probability parameter of the corresponding binomial) below the 0.15 used here. If we limit our interest to, say, the weight 1 and 2 perturbations, their number grows only quadratically with target size, while the overall number of perturbations grows exponentially: so with a fixed binomial coefficient, their influence on the fitness function declines rapidly. It may be possible, by suitably scaling the perturbation rate, to preserve their influence sufficently to realistically work with somewhat larger targets. 

In light of our observations regarding shadows of matrix~\ref{eq:patterned_2_grn_1}, if we restrict search to the subspace of modular solutions, the extended-target problem of~\citet{espinosa2010specialization} is highly reminiscent of the concatenated deceptive functions problem defined by~\citet{goldberg:cplxsys89}. It is generally accepted that effective general solutions to concatenated deceptive functions require some form of linkage learning. It is quite likely that algorithms such as the Messy Genetic Algorithm of ~\citet{goldberg:cplxsys89} or the Extended Compact Genetic Algorithm of~\citet{Illigal99010} would effectively solve these extended-target problems (even when not restricted to modular domains), and may promote modularity. However in this context, they require population-wide accumulation of information, and thus lack  biological plausibility. Progress in this direction would thus require both a biologically plausible algorithm accumulating and using linkage information, and a demonstrably feasible account ot how it might emerge. In this context, the crossover hotspot algorithm of~\citet{larson2016recombination} might provide useful hints.

\printbibliography
\end{document}